\documentclass[times, review, 10pt]{elsarticle}
\usepackage{geometry}
\usepackage{amssymb}
\usepackage{subfigure}
\usepackage{longtable}
\usepackage{multirow}
\usepackage{mathrsfs}
\usepackage{amsfonts}
\usepackage{amsmath}
\usepackage{array}
\usepackage{color}
\usepackage[ruled]{algorithm2e}
\usepackage{float}
\biboptions{numbers,sort&compress}

\journal{Pattern Recognition}

\begin{document}

\begin{frontmatter}



\title{Multi-Head Attention Driven Dynamic Visual-Semantic Embedding for Enhanced Image-Text Matching}


\author[firstauthor]{Wenjing Chen}

\cortext[cor1]{Email address: chenwj@stu.zuel.edu.cn (Wenjing Chen)}
\address[firstauthor]{School of Information Engineering, Zhongnan University of Economics and Law, Wuhan, 430073, China}

\begin{abstract}

With the rapid development of multimodal learning, the image-text matching task, as a bridge connecting vision and language, has become increasingly important. Based on existing research, this study proposes an innovative visual semantic embedding model, Multi-Headed Consensus-Aware Visual-Semantic Embedding (MH-CVSE). This model introduces a multi-head self-attention mechanism based on the consensus-aware visual semantic embedding model (CVSE) to capture information in multiple subspaces in parallel, significantly enhancing the model's ability to understand and represent the complex relationship between images and texts. In addition, we adopt a parameterized feature fusion strategy to flexibly integrate feature information at different levels, further improving the model's expressive power. In terms of loss function design, the MH-CVSE model adopts a dynamic weight adjustment strategy to dynamically adjust the weight according to the loss value itself, so that the model can better balance the contribution of different loss terms during training. At the same time, we introduce a cosine annealing learning rate strategy to help the model converge more stably in the later stages of training. Extensive experimental verification on the Flickr30k dataset shows that the MH-CVSE model achieves better performance than previous methods in both bidirectional image and text retrieval tasks, fully demonstrating its effectiveness and superiority.

\end{abstract}


\end{frontmatter}


\section{Introduction}
Image-text matching is one of the core tasks in the field of multimodal learning, and its importance is becoming increasingly prominent. This task aims to explore the intrinsic connection between vision and language, and has a far-reaching impact on promoting the application of artificial intelligence technology in multiple fields such as image annotation, image retrieval, and video understanding . In recent years, with the rapid development of deep learning technology, image-text matching methods have made significant progress\cite{antol2015vqa}. However, existing methods still face many challenges, especially in terms of in-depth exploration of the deep relationship between images and texts, which seriously limits the reasoning ability and matching accuracy of the model\cite{lee2018stacked}.

Traditional image-text matching methods mainly rely on global feature extraction, mapping images and texts to the same semantic space, and matching by calculating the distance or similarity between feature vectors. However, such methods often ignore the fine-grained information within the image and text, resulting in limited matching accuracy. With the rise of deep learning technology, image-text matching methods based on deep learning have gradually become mainstream\cite{wang2020consensus}\cite{zheng2020dual}. These methods can automatically learn high-dimensional feature representations of images and texts by training deep neural networks, thereby significantly improving matching performance.
Despite this, existing deep learning models still face challenges when dealing with complex scenes.

In particular, existing methods are often unable to capture the deep semantic relationship between images and text. To overcome this problem, this study proposed a new model called MH-CVSE (Multi-Headed Consensus-Aware Visual-Semantic Embedding). This model improves and innovates on the basis of the consensus-aware visual semantic embedding model (CVSE), introducing a multi-head self-attention mechanism, aiming to deeply explore the semantic connection between vision and text, thereby improving the performance of image-text matching\cite{wang2020consensus}.
 
The core idea of the MH-CVSE model is to capture the information of images and texts in parallel in multiple subspaces through a multi-head attention mechanism, thereby enhancing the model's ability to understand and represent complex relationships\cite{zheng2020dual}. In addition, the model also adopts a parameterized feature fusion strategy that can flexibly integrate feature information at different levels to further enhance the model's expressive power. In terms of loss function design, the MH-CVSE model innovatively adopts a dynamic weight adjustment strategy to dynamically adjust the weights according to the loss value itself, so that the model can better balance the contribution of different loss terms during training\cite{bruna2013spectral}. At the same time, in order to optimize the model's training process, this study also introduced a cosine annealing learning rate strategy to help the model achieve more stable convergence in the later stages of training.

\section{Related Work}

\subsection{Self-Attention Mechanism and Multi-Head Self-Attention}

The self-attention mechanism, especially the multi-head self-attention (MHSA) in the Transformer architecture, has become a core component in the field of natural language processing. Its powerful parallel processing capability and global information capture ability have enabled it to achieve remarkable results in multiple tasks\cite{chen2018temporally}. In the field of image-text matching, the self-attention mechanism has also shown great potential. Through the self-attention mechanism, the model can dynamically focus on the key information in the input image and text, thereby more accurately understanding the semantic relationship between the two.

This study extends the self-attention mechanism and introduces a multi-headed self-attention mechanism. The multi-headed self-attention mechanism transforms the query, key, and value by using multiple sets of linear projections in parallel, and performs attention calculations separately. The outputs of multiple attention heads are then concatenated together and transformed through another linear projection to produce the final output\cite{chen2018knowledge}. This mechanism allows the model to capture different aspects of the input data in different representation subspaces, thereby enhancing the model's ability to understand and represent cross-modal data\cite{chen2019multi}.

\subsection{Exploration and improvement of feature fusion technology}

Feature fusion is another key issue in multimodal learning, which involves how to effectively integrate feature information from different modalities. In the image-text matching task, the quality of feature fusion directly determines the matching performance of the model. Previous work has explored a variety of feature fusion methods, such as simple weighted summation, concatenation operations, and more complex dynamic weight fusion\cite{hou2022guidedstyle}. These methods have improved the matching accuracy to a certain extent, but there are still some limitations, such as the inability to fully capture the deep semantic relationship between cross-modal data\cite{engilberge2018finding}.

This study proposes a parameterized feature fusion method that allows the model to adaptively learn how to combine different features during training. Specifically, we design a trainable parameter matrix to weight and fuse features from different modalities. By optimizing these parameters, the model is able to learn the optimal feature combination, thereby achieving more flexible and effective feature fusion. This method not only improves the matching accuracy, but also enhances the generalization ability of the model\cite{lee2018stacked}.

\subsection{Dynamic Weight Adjustment of Loss Function and Learning Rate Scheduling Strategy}

In deep learning, loss function and learning rate are two crucial hyperparameters. The loss function is used to measure the difference between the models predicted results and the actual results, while the learning rate determines the step size of the models parameter update during training.

The loss function weights in the original model are fixed, which may result in the model being unable to effectively balance the contributions of different loss terms during training\cite{feng2024audios}. To solve this problem, this study implements dynamic adjustment of the loss function weights. Specifically, we dynamically adjust the weights of each loss term based on its value, so that the model can better balance the contributions of different loss terms during training. This approach helps improve the training efficiency and final performance of the model\cite{lin2014microsoft}.

In addition, the choice of learning rate has an important impact on the convergence speed and final performance of the model. Traditional fixed learning rates or simple decay strategies may not be able to adapt to complex training processes. To solve this problem, this study adopted the cosine annealing learning rate scheduling strategy\cite{krizhevsky2017imagenet}. This strategy dynamically adjusts the learning rate by simulating the annealing process of the cosine function, so that the model can converge more smoothly during training. Specifically, the learning rate is reset to a larger value at the beginning of each training cycle and gradually decreases to a smaller value during training. This strategy helps the model converge quickly in the early stages of training and maintain stable performance in the later stages of training. By adopting the cosine annealing learning rate scheduling strategy, this study further optimized the model's training process and improved the model's performance and stability\cite{kiros2014unifying}.

\section{MH-CVSE model}

The MH-CVSE model, or Multi-Headed Consensus-Aware Visual-Semantic Embedding model, is a model that enhances the understanding and representation of complex relationships between images and text by deepening the consensus-aware visual semantic embedding model (CVSE) and incorporating a multi-head self-attention mechanism\cite{wang2020consensus}. The model combines an image encoder and a text encoder, based on the Faster R-CNN network and a bidirectional GRU network, respectively, to extract and represent deep features of image regions and text sequences. The core multi-head self-attention module allows the model to process information in multiple subspaces in parallel, capturing richer contextual information through the transformation and attention calculation of queries (Q), keys (K), and values (V). In addition, the parameterized feature fusion module flexibly integrates feature information at different levels through the concat, $adap\_sum$, and $weight\_sum$ methods to further enhance the model's expressive power. The consensus-level feature learning module is built on top of a concept-related graph and uses a graph convolutional network to propagate semantic relationships between concepts to learn consensus-aware concept representations. During the training process, the dynamic loss weight adjustment mechanism dynamically adjusts the weights according to the loss value itself, so that the model can better balance the contribution of different loss terms. These components together constitute the architecture of the MH-CVSE model, enabling it to demonstrate superior performance in image-text matching tasks\cite{wang2020consensus}. The core architecture of the model includes the following key components:

\subsection{Model Architecture}

The architecture of the MH-CVSE model is shown in Figure 1 , which mainly includes the following modules:

\begin{figure}[!t]
	\begin{center}
		\includegraphics[width=0.8\linewidth]{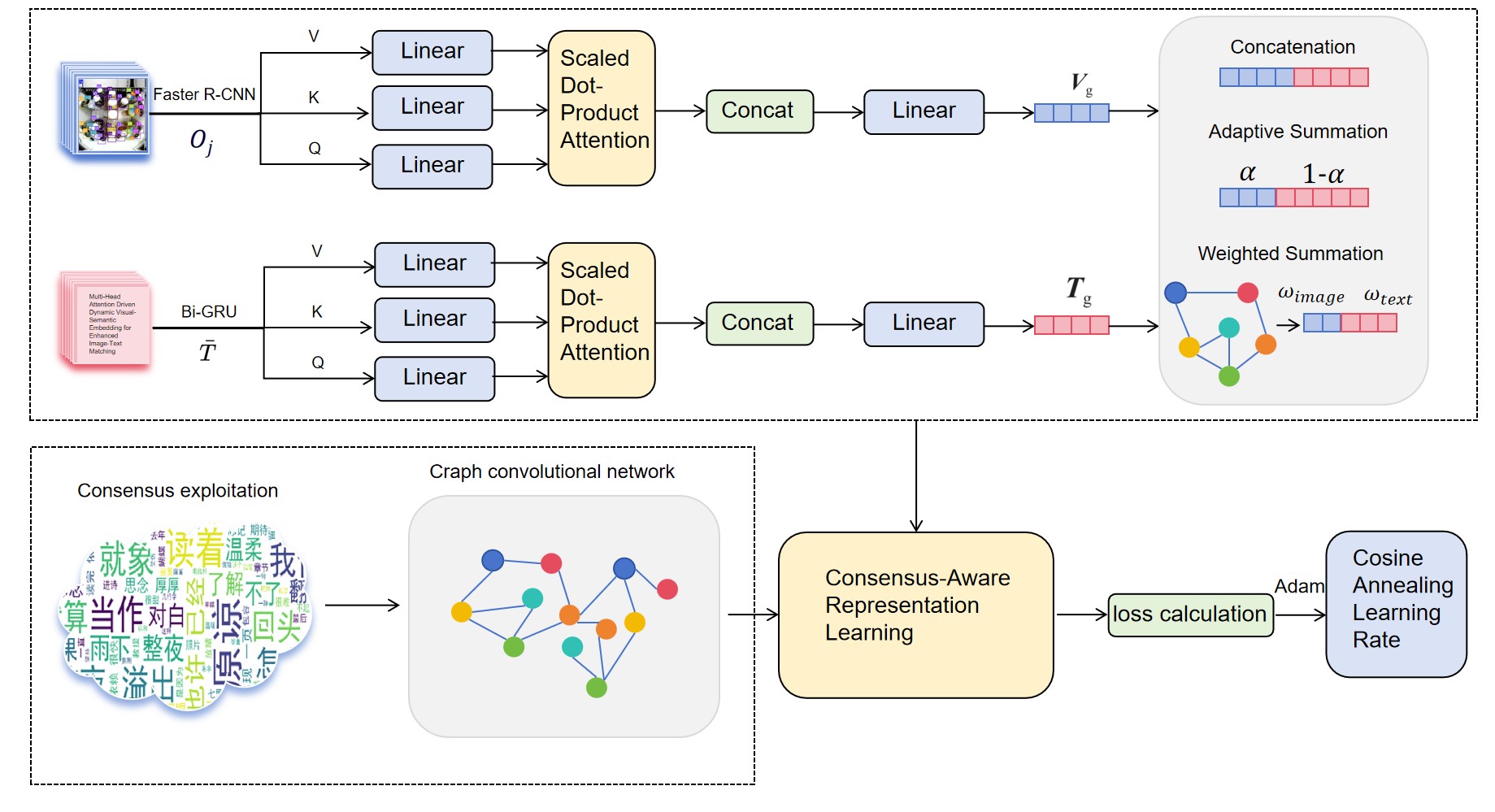}
	\end{center}
	\caption{MH-CVSE model integrates Faster R-CNN and Bi-GRU encoders with a multi-head self-attention mechanism for parallel image-text feature extraction. It employs parametric feature fusion (concat, $adap\_sum$, $weight\_sum$) and graph convolutional networks for consensus-aware learning. The model is further optimized with dynamic loss weights and cosine annealing learning rates for enhanced image-text matching performance.
	}
	\label{model4}
\end{figure}

Image encoder: The Faster R-CNN network based on deep learning is used as the image encoder [18] . The network generates candidate regions through the Region Proposal Network (RPN) and uses a convolutional neural network to extract the features of each region. The input is the original image and the output is the feature vectors of multiple regions, represented as:
\begin{equation}
	\mathbf{O} = \{\mathbf{o}_1, \mathbf{o}_2, \ldots, \mathbf{o}_M\}
\end{equation}where $M$ is the number of detected regions and the dimension of the feature vector is $F$.

Text encoder: The text encoder uses a bidirectional GRU (Bi-GRU) network\cite{anderson2017bottom}. The input is a text sequence processed by the embedding layer, and the output is the feature representation of the text. Specifically, the text sequence is converted into a word vector by the embedding layer and then input into the GRU. After forward and backward propagation, the output hidden state is average pooled to obtain the global representation of the sentence, denoted as T.

Multi-head self-attention module: This module introduces a multi-head self-attention mechanism, which allows the model to focus on multiple feature subspaces simultaneously when processing image and text data. The specific implementation is:
\begin{equation}
	\text{Attention}(Q, K, V) = \text{softmax}\left(\frac{QK^T}{\sqrt{d_k}}\right)V
\end{equation}Where $Q$, $K$, $V$ are the matrices of query, key, and value, respectively, and $ {d_k}$ is the dimension of the key. Through parallel computing of multiple heads, the model can capture richer contextual information and enhance the understanding of the complex relationship between images and text.

Parametric feature fusion module: This module integrates the feature representations of images and text by introducing methods such as $concat$, $adap\_sum$ and $weight\_sum$. The specific operations are as follows:

$concat$: directly concatenate the feature vectors of the image and text to form a new feature vector.

$adap\_sum$: Dynamically adjust weights based on data and calculate weighted sums.

$weight\_sum$: Uses an adaptive weight generator network to learn the importance of different features .

Consensus-level feature learning module: learns consensus-aware concept representation by building a concept-related graph. This module uses a graph convolutional network (GCN) to propagate the semantic relationship between concepts. The specific process is as follows:
\begin{equation}
	H^{(l+1)} = \sigma(A \cdot H^{(l)}) \cdot W^{(l)}
\end{equation}Where $H^{(l)}$ is the point feature matrix of the l-th layer node , $A$ is the normalized adjacency matrix, $ W^{(l)}$ is the learnable weight matrix, $ \sigma$ and is the activation function.

Dynamic loss weight adjustment mechanism: During the training process, the model dynamically adjusts the weights of different loss items according to the loss value. The specific implementation is:
\begin{equation}
	L = \sum_{i} \lambda_i L_i
\end{equation}
Where $ L_i $ is the i- th loss term and $\lambda_i $ is the dynamically adjusted weight. This design enables the model to better balance the contributions of different loss terms during training, improving the robustness and generalization ability of the model.

\subsection{Multi-head self-attention mechanism}

The MH-CVSE model enhances the model's ability to capture information from different subspaces by introducing a multi-head self-attention mechanism. The multi-head self-attention mechanism is an effective technique that allows the model to focus on multiple feature subspaces simultaneously when processing image and text data, thereby capturing richer contextual information. Specifically, the mechanism is implemented through the following steps , as shown in Figure 2 :

\begin{figure}[!t]
	\begin{center}
		\includegraphics[width=0.8\linewidth]{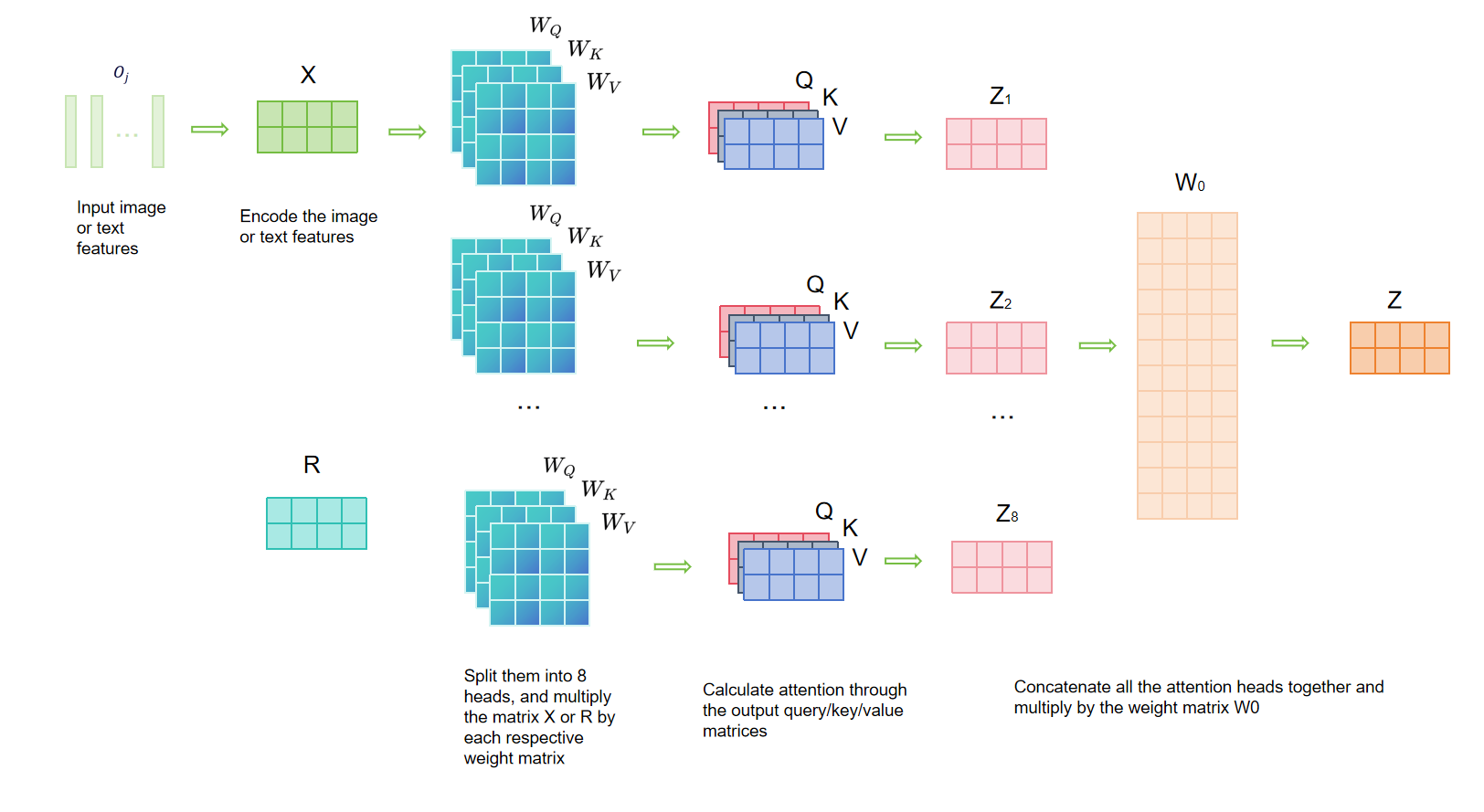}
	\end{center}
	\caption{Schematic of the multi-head self-attention mechanism. The process begins with encoding the input image or text features (X), which are then split into multiple attention heads. Each head computes attention using separate weight matrices ($W_Q$, $W_K$, $W_V$) for queries, keys, and values, producing outputs ($Z_1$, $Z_2$, ...,$Z_8$). These outputs are concatenated and transformed by a final weight matrix ($W_O$) to yield the final attention output (Z).
	}
	\label{multihead2}
\end{figure}

Define query, key, and value: For the input feature matrix X, we generate query Q, key K, and value V through linear transformation:
\begin{equation}
	Q = XW_Q
\end{equation}
\begin{equation}
	K = XW_K
\end{equation}
\begin{equation}
	V = XW_V
\end{equation}
Among them, $ W_Q$, $ W_K$ and $ W_V$ are learnable parameter matrices.

Calculate the attention score: Calculate the correlation between the query and the key through the dot product to get the attention score:
\begin{equation}
	\text{score}(Q, K) = \frac{Q^T K}{\sqrt{d_k}}
\end{equation}
Where $d_k$ is the dimension of the key, and is scaled using a square root $\sqrt{d_k}$ to avoid large values when calculating the softmax.

Apply the softmax function: perform softmax normalization on the attention score to obtain the attention weight:
\begin{equation}
	\text{Attention}(Q, K, V) = \text{softmax}\left(\frac{QK^T}{\sqrt{d_k}}\right)V
\end{equation}

Multi-head attention calculation: In the multi-head attention mechanism, we use hh groups of independent linear transformations to generate different queries, keys, and values. The outputs of each group of attention heads are concatenated and then linearly transformed to obtain the final output:
\begin{equation}
	\text{MultiHead}(Q, K, V) = \text{Concat}(\text{head}_1, \ldots, \text{head}_h)W^O
\end{equation}

The calculation for each head is:
\begin{equation}
	\text{head}_i = \text{Attention}(QW_i^Q, KW_i^K, VW_i^V)
\end{equation}
$ W^O$ is the output linear transformation matrix.
This mechanism not only enhances the model's ability to capture important information in images and texts, but also promotes the effective fusion of cross-modal information, thereby improving the accuracy of matching. Through the multi-head self-attention mechanism, the MH-CVSE model can better understand the complex relationship between images and texts, thereby improving the performance of image-text matching.

\subsection{Parameterized Feature Fusion}

In the CVSE model proposed in this study, we adopted a parameterized feature fusion strategy to enhance the model's ability to integrate image and text features. Specifically, we implemented three feature fusion methods: concatenation (Concat), adaptive weighted summation ($Adap\_sum$), and weight-based weighted summation ($Weight\_sum$). These methods are implemented through the $Multi\_feature\_fusing$ module, which dynamically selects the fusion strategy based on the input $fuse\_type$ parameter. In practical applications, we can flexibly choose the most appropriate feature fusion method based on data characteristics and task requirements to achieve the best model performance. The following are the three feature fusion methods we implemented:

Concatenation

Splicing is the simplest feature fusion method. It directly connects the feature vectors of the image and text to form a new feature representation\cite{chen2018temporally}. This method is simple and direct, can retain all the information of the original features, and provide the model with more feature dimensions for learning.
\begin{equation}
	V_{\text{concat}} = [V_{\text{image}}; V_{\text{text}}]
\end{equation}
Among them, $V_{\text{image}}$ is the image feature vector, $V_{\text{text}}$ is the text feature vector, and the concatenated feature vector $V_{\text{concat}}$ contains all the information of the image and text.

Adaptive Sum

The adaptive weighted sum method dynamically adjusts the weights of image and text features according to the characteristics of the input data. Through the learned weights, the model can better combine information from different modalities.
\begin{equation}
	V_{\text{adap\_sum}} = \alpha \cdot v_{\text{image}} + (1 - \alpha) \cdot v_{\text{text}}
\end{equation}
Among them, $\alpha $ is a learnable weight that represents the importance of image features.

The weighted summation method based on weights allows the model to learn the importance of different features. Through an adaptive weight generation network, the model can assign a weight to each feature, thereby achieving more effective information integration.
\begin{equation}
	v_{\text{weight\_sum}} = w_{\text{image}} \cdot v_{\text{image}} + w_{\text{text}} \cdot v_{\text{text}}
\end{equation}
Among them, $ w_{\text{image}}$ and $ w_{\text{text}}$ are weights learned through training, indicating the importance of image and text features.

This parameterized feature fusion strategy not only improves the flexibility of the model, but also enhances its ability to integrate information from different modalities, thereby achieving significant performance improvements in image-text matching tasks.

\subsection{Dynamic Weight Adjustment of Loss Function}

In the training process of deep learning models, the design of the loss function has a crucial impact on the model performance. In traditional models, the weights of the loss function are usually pre-set and fixed, which may cause the model to over-rely on certain loss terms during training and ignore other important loss terms. To address this problem, our MH-CVSE model introduces an innovative dynamic adjustment strategy for the loss function weights, which can adaptively adjust the weights according to the performance of the model during training to achieve a more balanced loss contribution.

In our method, the loss function consists of four parts: instance-level contrastive loss, consensus-level contrastive loss, fusion-level contrastive loss, and KL divergence loss. The weights of these loss terms are not fixed, but dynamically adjusted according to the current loss value. Specifically, we use an exponential decay-based function to adjust the weight of each loss term, as follows:
\begin{equation}
	\text{loss\_weight} = w \times \frac{1}{1 + e^{-\text{loss\_value}}}
\end{equation}
Among them, \( w \) is the pre-set basic weight, and \( \text{loss\_value} \) is the current value of the corresponding loss item.This adjustment mechanism makes it so that when the value of a loss item is high, its weight will be reduced accordingly, and vice versa. This design helps the model not to pay too much attention to a specific loss item during training, but to consider all loss items more evenly, thereby improving the training efficiency and generalization ability of the model.

In our implementation, the base weights are set to a value optimized for the F30k dataset. These weights are determined experimentally to ensure that the various loss terms can complement each other and jointly drive model performance improvements at different stages of training.Ultimately, the total loss is the sum of all weighted loss terms.

Through this dynamic weight adjustment strategy, the MH-CVSE model can adaptively adjust the importance of different loss terms during the training process, so that the model not only performs well on the training set, but also shows strong generalization ability on unseen test data. The application of this strategy provides important support for our model to achieve excellent performance in image-text matching tasks.

\subsection{Cosine Annealing Learning Rate Strategy}
In order to further improve the training efficiency and stability of the MH-CVSE model, we introduced the cosine annealing learning rate strategy. This strategy dynamically adjusts the learning rate by simulating the periodic changes of the cosine function, thereby providing a larger learning rate in the early stage of training to speed up convergence, and gradually reducing the learning rate in the later stage of training to achieve more precise parameter adjustment. Specifically, the learning rate is adjusted according to the following formula:
\begin{equation}
	\eta_t = \eta_0 \times \left( \text{annealing function of } t \text{ and } T \right)
\end{equation}
Where $\eta_t $ is the learning rate of the tth training step , $\eta_0$ is the initial learning rate, $T$ is the set annealing period, and t is the current training step.

This learning rate adjustment strategy helps the model converge more stably in the later stage of training, avoids overfitting, and improves the generalization ability of the model. In the MH-CVSE model, the application of the cosine annealing learning rate strategy enables the model to achieve better performance on the Flickr30k dataset, which is verified in the experimental part.

\subsection{Training and Inference}

The training phase of the MH-CVSE model uses the Adam optimizer combined with dynamic loss weight adjustment and a cosine annealing learning rate strategy. During the training process, the model first extracts features through the image encoder and text encoder, and then integrates these features through the self-attention module and the feature fusion module. Subsequently, the model is optimized using a dynamically adjusted loss function, which includes contrastive loss and KL divergence loss. This training method enables the model to adaptively adjust the weights of different loss terms and dynamically adjust the learning rate, thereby achieving more efficient parameter updates during the training process. In this way, the MH-CVSE model is able to achieve good performance on the training set and show strong generalization capabilities on unseen test data.

In the inference phase, the MH-CVSE model uses the trained parameters to match new image-text pairs. The model achieves fast and accurate matching by computing the embedded representations of the image and text and calculating the distance between them in the joint embedding space. Specifically, the model calculates the similarity of image and text features and uses the cosine similarity metric to ensure efficient execution on large-scale datasets. This approach not only improves the accuracy of matching, but also ensures the efficiency of the model in practical applications. Experimental results on the Flickr30k dataset show that the MH-CVSE model achieves better performance than previous methods in both bidirectional image and text retrieval tasks, fully demonstrating its effectiveness and superiority.

\section{Experiment}

\subsection{Dataset and Settings}

The experiments used the widely used Flickr30k dataset, which contains 31,783 images, each with 5 descriptive texts , as shown in Figure 3. We followed the standard dataset partitioning method, using 29,783 images for training, 1,000 for validation, and 1,000 for testing. We focused on the image-to-text retrieval task and evaluated the model performance on the test set.

\begin{figure}[!t]
	\begin{center}		\includegraphics[width=0.8\linewidth]{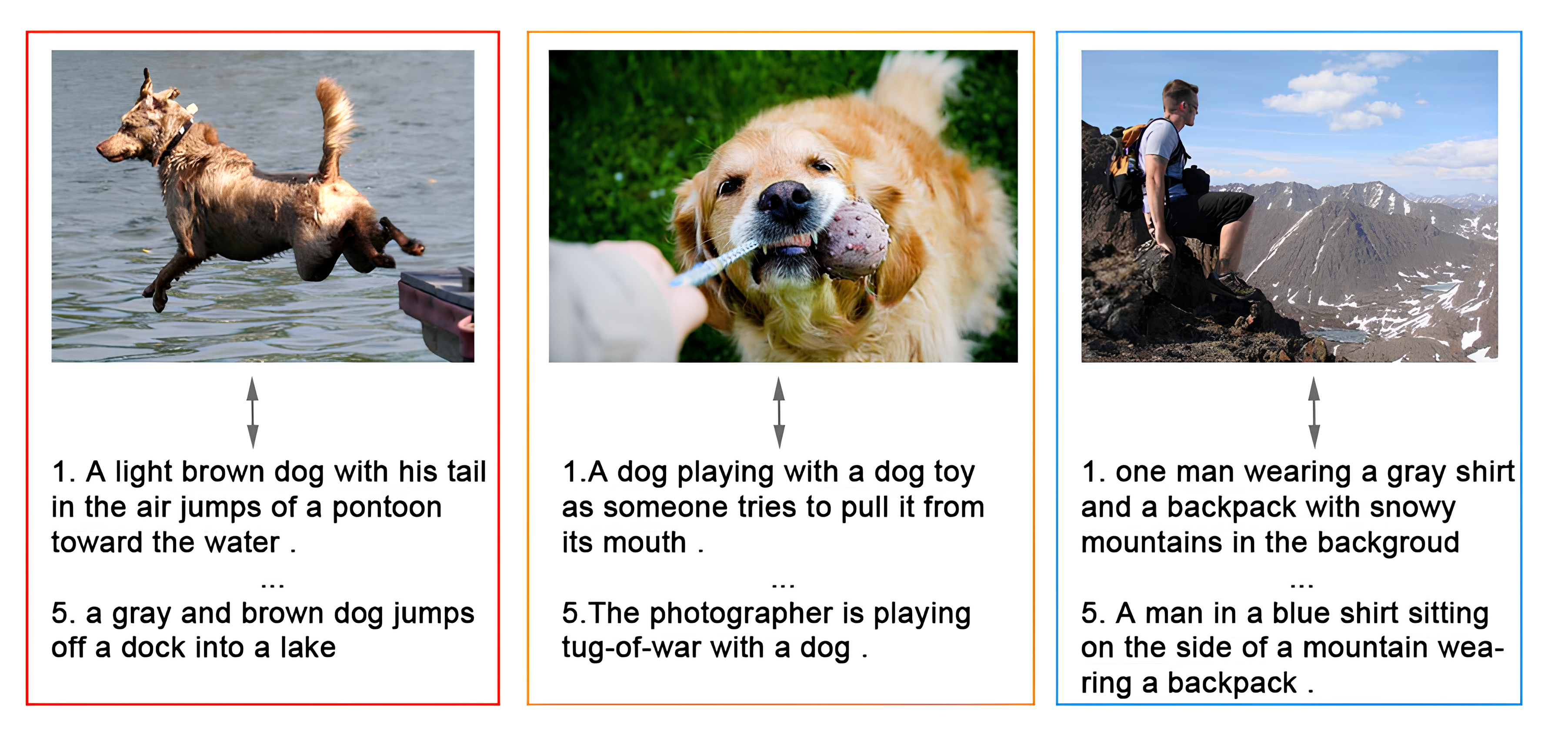}
	\end{center}
	\caption{ Image-text matching examples. From left to right: (1) a dog leaping off a dock, (2) a dog in a tug-of-war with a toy, (3) a man with a backpack on a mountain trail. Each image is paired with a textual description highlighting the main action and elements\cite{zheng2020dual}.
	}
	\label{datasets}
\end{figure}

\subsection{ Evaluation Metrics}

We adopt commonly used evaluation metrics in image-text matching tasks, including Recall@K (R@K), which measures the proportion of relevant terms in the top K retrieval results for each query in the retrieval task. We report the results of R@1, R@5, and R@10, as well as the mean recall (mR)\cite{wang2020consensus}.

\subsection{Experimental Details}

During the experiment phase of the MH-CVSE model, we adopted a rigorous setup to ensure the reproducibility and transparency of the experimental results. All experiments were conducted on a server equipped with an NVIDIA RTX 3080 GPU, using PyTorch 1.8.1, Python 3.8, and CUDA 11.1. The Flickr30k dataset was carefully preprocessed, the images were resized and normalized to fit the model input requirements, and the text data was converted into word embeddings after tokenization and stop word removal. The model configuration includes a Faster R-CNN image encoder and a bidirectional GRU (Bi-GRU) text encoder, each followed by an 8-head self-attention module for further feature extraction. In the feature fusion strategy, we adopted the weighted sum method because it showed the best performance in the experiment. Consensus-level feature learning is implemented through a graph convolutional network (GCN) to learn the semantic relations between concepts. The model was trained for 30 epochs, with performance evaluation performed after each epoch\cite{wang2020consensus}. The early stopping strategy was used to stop training when there was no performance improvement on the validation set for 5 consecutive epochs. The evaluation was performed on the Flickr30k test set, focusing on the image-to-text retrieval task. In the experiment, we paid special attention to the implementation details of the model components to ensure computational efficiency and model performance.

\subsection{Comparative Experiment}

In order to verify the effectiveness of each component in the model, we conducted a comparative experiment. The results of the comparative experiment are shown in Table 1. We can see that after introducing the multi-head self-attention mechanism, parameterized feature fusion, dynamic loss weight adjustment, and cosine annealing learning rate strategy, the model performance has steadily improved. In particular, when all four improvements are combined, the MH-CVSE model achieves the best performance, as shown in Table 1.

\begin{table*}
	\caption{Performance comparison on the Flickr30k dataset.}
	\begin{center}
		\footnotesize
		\begin{tabular}{lccccccc}
			\hline
			Approach & \multicolumn{3}{c}{Text retrieval} & \multicolumn{3}{c}{Image Retrieval} & mxD \\ \hline
			& R@1 & R@5 & R@10 & R@1 & R@5 & R@10 &  \\ \hline
			DVSA & 22.2 & 48.2 & 61.4 & 15.2 & 37.7 & 50.5 & 58.5 \\
			m-RNN & 35.4 & 63.8 & 73.7 & 22.8 & 50.7 & 63.1 & 51.6 \\
			DSPE & 40.3 & 68.9 & 79.9 & 29.7 & 60.1 & 72.1 & 58.5 \\
			CMPM & 49.6 & 76.8 & 86.1 & 37.3 & 65.7 & 75.5 & 65.2 \\
			VSE++ & 52.9 & - & 87.2 & 39.6 & - & 79.5 & - \\
			PVSE & - & - & - & - & - & - & - \\
			SCAN & 67.4 & 90.3 & \textbf{95.8} & 48.6 & 77.7 & 85.2 & 77.5 \\
			CAMP & 68.1 & 89.7 & 95.2 & 51.5 & 77.1 & 85.3 & 77.8 \\
			LIWE & 69.6 & 90.3 & 95.6 & 51.2 & 80.4 & 87.2 & 79.1 \\
			CVSE & 73.5 & 92.1 & 95.8 & 52.9 & 80.4 & 87.8 & 80.4 \\
			MH-CVSE & \textbf{74.1} & \textbf{92.6} & {95.4} & \textbf{53.8} & \textbf{82.1} & \textbf{88.9} & \textbf{83.5} \\ \hline
		\end{tabular}
	\end{center}
	\label{tabflickr30k}
\end{table*}

\subsection{Results Analysis}

As shown in Table 1, the MH-CVSE model shows excellent performance in both text retrieval and image retrieval tasks. In the text retrieval task, MH-CVSE achieved an R@1 accuracy of 74.1\%, an R@5 accuracy of 92.6\%, and an R@10 accuracy of 95.4\%, significantly surpassing all other comparison methods. In the image retrieval task, MH-CVSE also achieved an R@1 accuracy of 53.8\%, which is the highest value among all methods, further proving the effectiveness of the model in bidirectional retrieval tasks.

\subsection{Visualization of experimental results}

To demonstrate the models image-text matching capability, we randomly select an image from the test set , as shown in Figure 4 , and use the model to calculate its similarity with all text descriptions in the dataset. The text with the highest similarity is selected as the matching result for visualization.
\begin{figure}[!t]
	\begin{center}
		\includegraphics[width=0.8\linewidth]{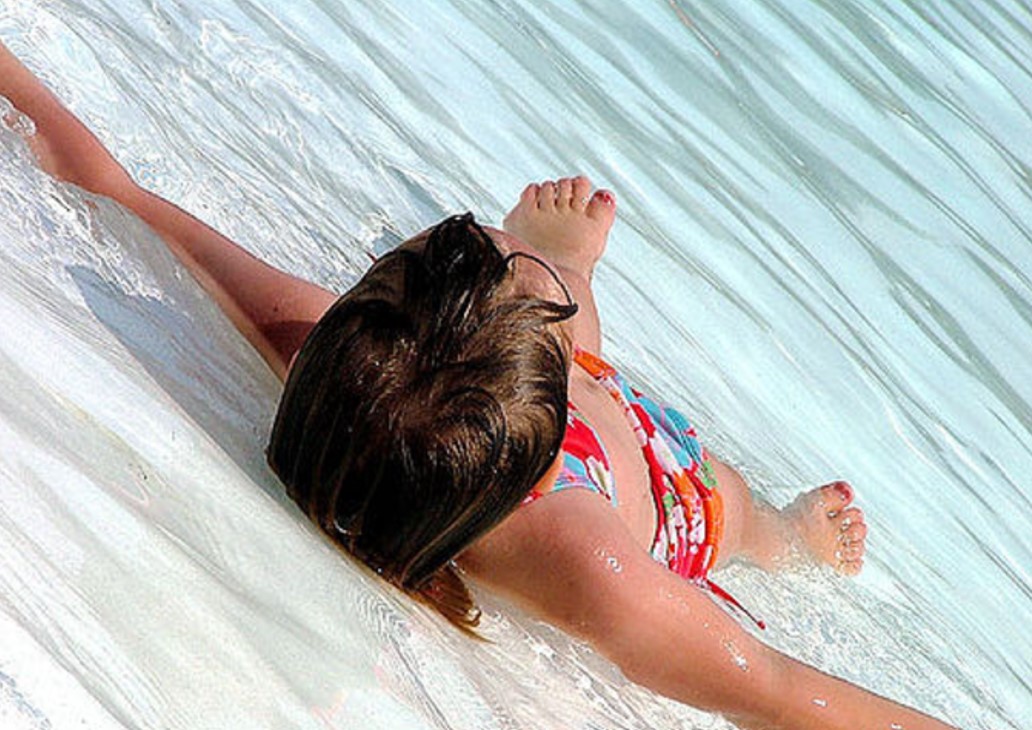}
	\end{center}
	\caption{ Matching text: A girl wearing a red and multicolored bikini is laying on her back in shallow water .
	}
	\label{result3}
\end{figure}

It is clear from the visualization results that the model successfully captures the key elements in the picture, such as the girl, the colorful swimsuit, floating on the water, and the outstretched limbs, and accurately reflects them in the text description. This shows that the model can effectively understand image content and match it with semantically related text.

The model integrates different levels of feature information from images and text through a parameterized feature fusion strategy. In this example, visual features from the image (such as the girl's posture, the color of the swimsuit, etc.) are well fused with semantic features from the text (such as the words "floating" "swimsuit" etc.), thereby achieving accurate matching.

The multi-head self-attention mechanism enables the model to capture information in multiple subspaces in parallel, focusing on different aspects of the image and text. For example, one attention head might focus on the girl's actions, while another might focus on the color of the swimsuit. These pieces of information work together to enhance the model's ability to understand and represent complex relationships, thus accurately matching the image with the text that describes its content.

\section{Conclusion}

This study successfully proposed an innovative image-text matching model, Multi-Headed Consensus-Aware Visual-Semantic Embedding (MH-CVSE). By introducing a multi-head self-attention mechanism based on the existing visual semantic embedding model (CVSE)\cite{wang2020consensus}, our model significantly improves the ability to understand and represent the complex relationship between images and text. In addition, the parameterized feature fusion strategy we adopted further enhances the model's ability to integrate feature information at different levels.

Extensive experiments on the Flickr30k dataset validate the effectiveness of the MH-CVSE model. Experimental results show that MH-CVSE outperforms existing state-of-the-art methods in both bidirectional image and text retrieval tasks. This fully demonstrates the effectiveness and superiority of our model in image-text matching tasks.

The main contributions of our work include:

Multi-head self-attention mechanism: For the first time, the multi-head self-attention mechanism is applied to the image-text matching task to effectively capture the deep semantic relationship of cross-modal data.

Parametric feature fusion strategy: By flexibly integrating feature information at different levels, the expressiveness of the model is further improved.

Dynamic loss weight adjustment: Dynamically adjust the weights according to the loss value itself, so that the model can better balance the contribution of different loss terms during training.

Cosine annealing learning rate strategy: Introduce the cosine annealing learning rate strategy to help the model converge more stably in the later stages of training.

\leftline{ {\bf Acknowledgements}} The completion of this study is inseparable from the hard work and valuable contributions of all authors. First of all, we would like to thank Haoran Wang, Ying Zhang, Zhong Ji, Yanwei Pang and Lin Ma, whose research results provided the basic code and implementation ideas for our model. In particular, Haoran Wang, his pioneering work on the CVSE model provided us with valuable reference and inspiration.

We would also like to thank Kuang-Huei Lee, Xi Chen, Gang Hua, Houdong Hu, and Xiaodong He for providing us with in-depth insights and rich dataset features on the image-text matching task through the Stacked Cross Attention for Image-Text Matching paper. We are especially grateful to Kuang-Huei Lee for his work on dataset feature extraction and model evaluation, which provided an important experimental basis and performance comparison for our research.



\quad



\bibliography{mybibfile}

\end{document}